\documentclass{article}
\usepackage{spconf,amsmath,graphicx}
\usepackage{multirow}
\usepackage{booktabs}
\usepackage[colorlinks = true,
            linkcolor = blue,
            urlcolor  = blue,
            citecolor = blue,
            anchorcolor = blue]{hyperref}

\newcommand{\MYhref}[3][blue]{\href{#2}{\color{#1}{#3}}}%

\usepackage{xcolor}

\title{Heart rate estimation in intense exercise videos}
\name{Y. Napolean$^1$*, A. Marwade$^1$*, N. Tomen$^1$, P. Alkemade$^2$, T. Eijsvogels$^3$, J.C. van Gemert$^1$\thanks{* Equal contribution}}
\address{TU Delft$^1$, VU Amsterdam$^2$, Radboud UMC$^3$}
\begin{document}
%
\maketitle
\vspace{-4mm}
\begin{abstract}
Estimating heart rate from video allows non-contact health monitoring with applications in patient care, human interaction, and sports. Existing work can robustly measure heart rate under some degree of motion by face tracking. However, this is not always possible in unconstrained settings, as the face might be occluded or even outside the camera. Here, we present IntensePhysio: a challenging video heart rate estimation dataset with realistic face occlusions, severe subject motion, and ample heart rate variation. To ensure heart rate variation in a realistic setting we record each subject for around 1-2 hours. The subject is exercising (at a moderate to high intensity) on a cycling ergometer with an attached video camera and is given no  instructions regarding positioning or movement. We have 11 subjects, and approximately 20 total hours of video. We show that the  existing remote photo-plethysmography methods  have difficulty in estimating heart rate in this setting. In addition, we present IBIS-CNN, a new baseline using spatio-temporal superpixels, which improves on existing models by eliminating the need for a visible face/face tracking. We will make the code and data publically available soon.\footnote{\MYhref{https://github.com/ynapolean/IBIS-CNN}{https://github.com/ynapolean/IBIS-CNN} }

\end{abstract}
\section{Introduction}
\label{sec:intro}

Remote photo-plethysmography (rPPG)~\cite{zhan2020analysis}, allows non-contact heart rate estimation. This facilitates applications where contact sensors are difficult such as  infant health  monitoring, or athlete monitoring in large scale events like marathons, where access to individual sensors is unavailable.  Initial methods of rPPG estimation \cite{kwon2012validation,verkruysse2008remote} needed the subject to sit motionless in front of a camera. Other approaches \cite{jonathan2010investigating, gudi2019efficient, gudi2020real, zhu2012non} typically used face detectors and Fourier analysis based techniques.
 Some methods additionally use skin tracking/segmentation to estimate heart rate \cite{tang2018non,yu2019remote}. Other approaches  cast heart rate estimation from video as a blind source separation problem~\cite{poh2010non,6078233}. Most existing rPPG methods including recent deep learning based approaches~\cite{niu2018vipl, chen2018deepphys} typically operate within a `constrained' setting: minor pose and camera angle variations, subjects are cooperative, are not occluded and do not perform high-speed movements. Enforcing such constraints is  not possible in real-life situations. For instance in a sports scenario, it is hard to control lighting conditions and rapid pose changes.
 
 \begin{figure}
    \centering
    \includegraphics[scale = 0.38,trim=0 20 0 5]{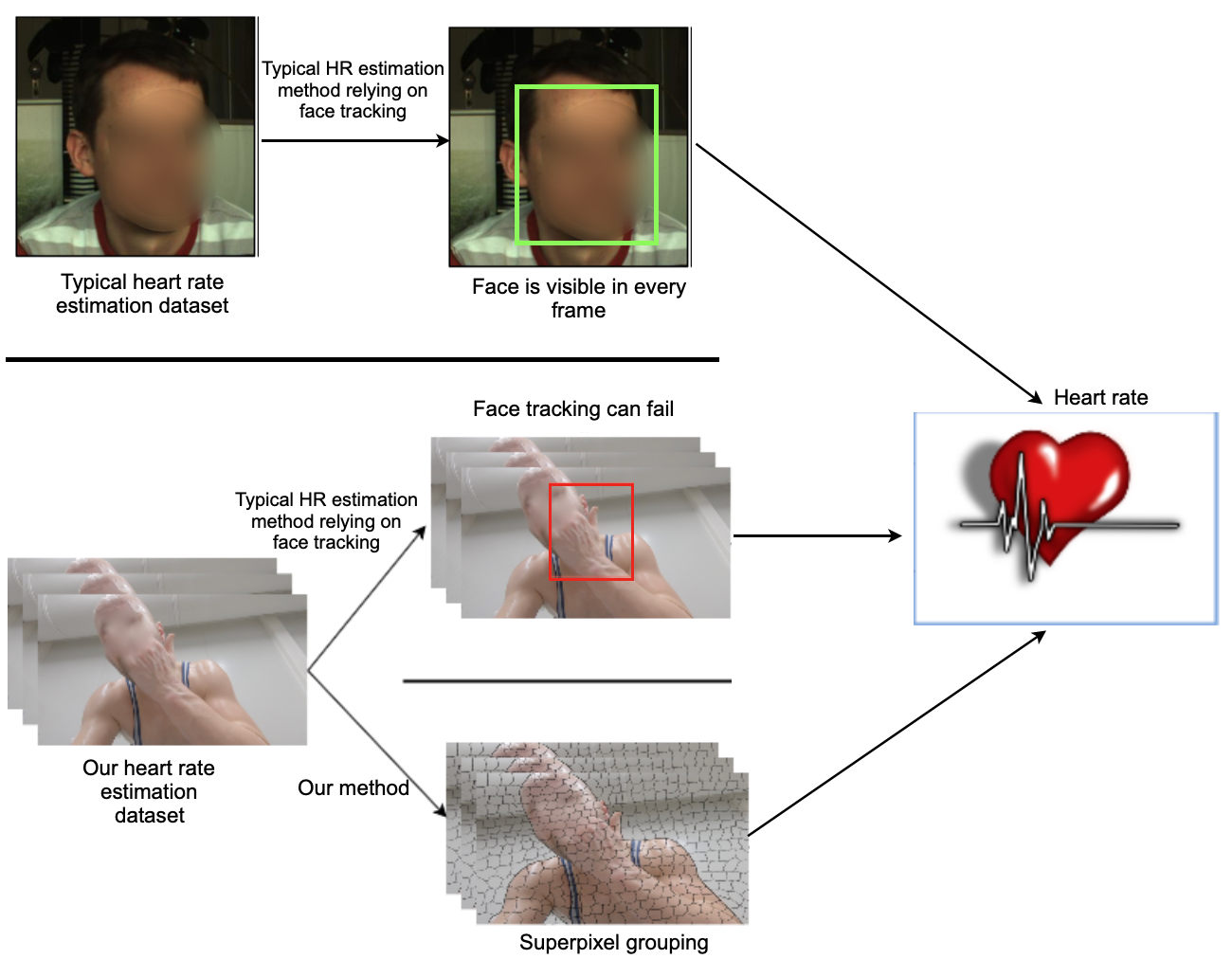}
    \caption{Methods tracking skin pixels and their color change for heart rate estimation can perform poorly when the subject's face is not visible. Here we introduce the IntensePhysio dataset, which includes high heart rate variability and is recorded in a relatively `unconstrained' setting, making it challenging for existing methods. We propose the IBIS-CNN method which relies on spatio-temporal grouping of pixels instead of face tracking, which improves on baseline models.\vspace{-3mm}}
    \label{fig:my_label}
\end{figure}

A recent analysis \cite{Zhao_2019_ICCV} shows that large, high speed motion,  tracking based methods can fail. There are methods that have been proposed to handle partial facial occlusion \cite{9176563}\cite{s20247021}, but these methods rely on extracting information from other parts of the face that are visible and do not address facial occlusion (as depicted in Fig. 1). To develop a method for heart rate estimation from video in  a practical setting (like sports, for example the last image from the left in the second row of Fig. 2), this must be addressed.


Early datasets, MAHNOB~\cite{soleymani2011multimodal} and COHFACE~\cite{heusch2017reproducible} feature almost no subject motion. 
The PURE dataset features minimal controlled head motion \cite{stricker2014non}. 
An important step towards more realistic scenarios is the seminal ECG-Fitness dataset~\cite{vspetlik2018visual}  where subjects perform fitness-related motions under varying lighting conditions, demonstrating that previous visual heart rate estimation methods performed poorly in this scenario. This ECG-Fitness dataset, however, assumes that faces are still strictly visible, allowing for solutions limited to face tracking.

In this work, we introduce IntensePhysio: a new, publicly available, challenging dataset with ample variance in the heart rate, fast motions, severe appearance changes. It presents a completely unconstrained setting wherein subjects are allowed exacerbated motion, causing their face to be sometimes occluded, or they can go out of frame and exhibit greatly varying face angles.
In addition to the new dataset, we analyze existing rPPG approaches and provide an intentionally simple baseline method called IBIS-CNN that replaces face tracking with spatio-temporal superpixels. Bobbia\cite{bobbiaibis} introduced an algorithm called Iterative Boundaries implicit Identification for superpixel Segmentation (IBIS)~\cite{bobbiaibis} for real-time computation of temporal superpixels for rPPG. We propose IBIS-CNN as a novel application of IBIS for heart rate estimation in a deep learning setting. IBIS-CNN performs on-par to existing methods on existing datasets, but significantly outperforms existing work on more realistic scenarios as exemplified in our new IntensePhysio dataset.
The contributions of this paper are: (i) The new, public, IntensePhysio dataset with facial occlusions, and high speed motion (ii) A simple baseline method using superpixels which does not rely on the face to extract heart rate. (iii) We demonstrate the difficulty of the dataset for existing rPPG work and (iv) show that  our simple baseline performs equally well on existing datasets, yet significantly better on our more realistic dataset.
\vspace{-0.3cm}

\section{Method}
\label{sec:Method}
\subsection{The IntensePhysio dataset}

\begin{figure}
\centering
\includegraphics[width=\linewidth,trim=0 0 0 20]{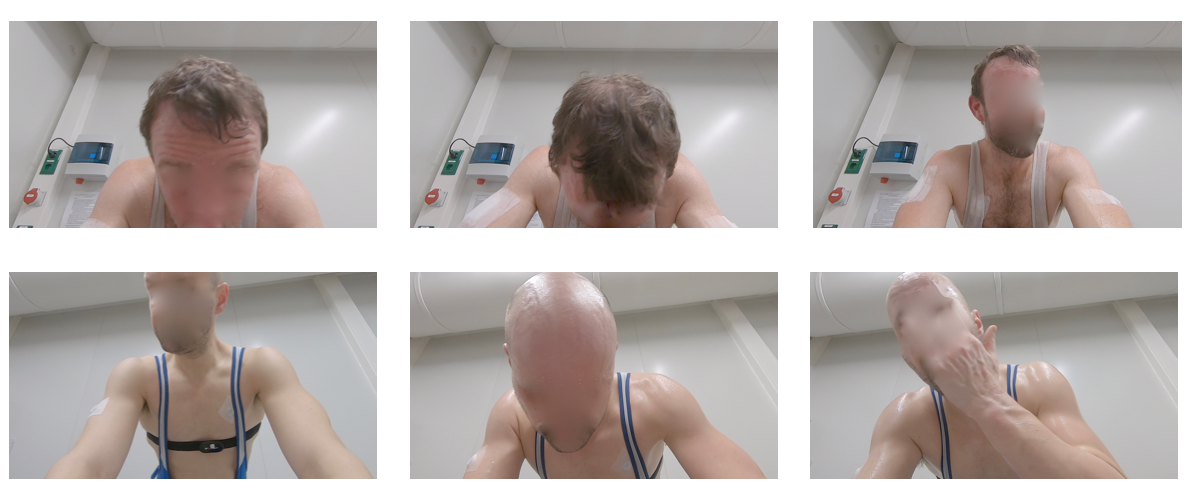}
\vspace{-0.8cm}
\caption{Samples from our IntensePhysio rPPG dataset. Note the major pose changes, specular reflections on the background and on the subject's skin and instances where the face region is occluded or not fully in frame.\vspace{-0.3cm}}
\label{fig:dataset}
\end{figure}

We collected video recordings and heart rate as the ground truth. The videos are recorded with a Go-Pro Hero 7 black camera at a resolution of $1,920 \times 1,080$ pixels at $\sim$ 60 frames per second. The ground truth heart rate (in beats per minute) is measured every second, using the Polar Vantage-M heart rate sensor (chest strap). 

For illumination we use LED lights (47W), natural light is also allowed to enter through a small window. Recordings occur during different times of the day, resulting in variation of lighting conditions. There are a total of 11 subjects (3 female and 8 male) and all participants are working out on an ergometer and are given no instructions except to do a workout. This dataset features the largest heart rate range and variation: from 51 bpm up to 186 bpm with a standard deviation of 25 bpm. IntensePhysio dataset statistics are presented in Table \ref{dataset_stats}. Subjects exercised at a moderate to high intensity. They did not receive any camera-related instructions, such as to face the camera or be in the frame. This means that the subject's motion is 'unconstrained' and consequently the subject's face  could be occluded or might not be visible at all, as illustrated in Fig.~\ref{fig:dataset}. 
\vspace{-0.4cm}

\begin{table}
\centering
\begin{tabular}{ll}
\toprule
\multicolumn{2}{c}{\textbf{Dataset Statistics}}     \\ \midrule
Mean heart rate (ground truth)           & 129         \\
Std. deviation heart rate (ground truth) & 25            \\
Max. heart rate (ground truth)                         & 186         \\
Min. heart rate (ground truth)                         & 51          \\
No. of subjects                          & 11                 \\
No. of videos                            & 15                 \\
Avg. video runtime                       & 1hr. 12min. \\
Frames per second                        & 59.94            \\
\bottomrule
\end{tabular}
\caption{Statistics of IntensePhysio. The dataset features a wide range of heart rates, from resting to intense workout. Heart rate values are in beats per minute.\vspace{-0.5cm}}
\label{dataset_stats}
\end{table}

\subsection{Simple rPPG baseline: IBIS-CNN}

\begin{figure*}[h!]
\centering
\vspace{-3mm}
\includegraphics[width=0.95\textwidth]{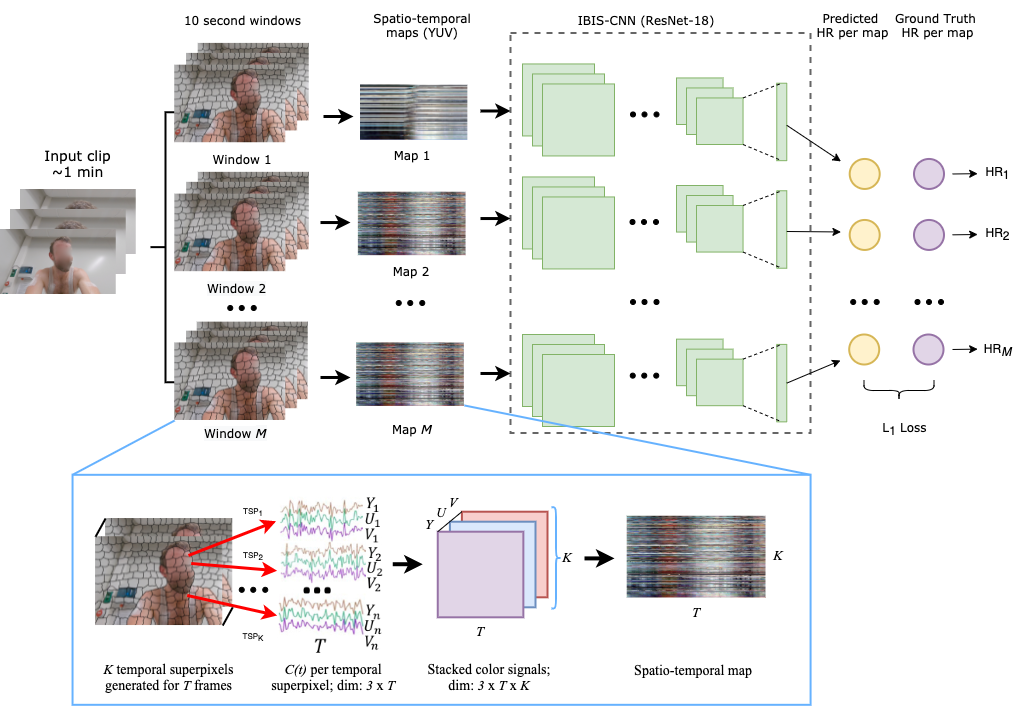}
\vspace{-3mm}
\caption{An overview of the IBIS-CNN model. A clip is split into $M$ windows of 10 seconds each. For each frame, $K$ superpixels are extracted and per superpixel, the average YUV color is stacked over time, where $T$ is the number of frames  and generate a spatio-temporal map. The predicted heart rate per map from the CNN are averaged to estimate the HR per video. \vspace{-2mm} }
\label{fig:model}
\end{figure*}

To accompany the IntensePhysio dataset, we implement a simple baseline method for rPPG estimation using temporal superpixels and a convolutional neural network. The approach is shown in Fig. \ref{fig:model}.

The model is trained to predict the scalar valued instantaneous heart rate in beats per minute (bpm) for an input video. We break an input video sequence into non-overlapping clips of one minute each. For each clip we generate $K$ temporal superpixels using IBIS~\cite{bobbiaibis} where $K$ refers to a user-defined value for the number of superpixels, which we set to 300. 


The input images are converted to the $CIE\;l^*a^*b^*$ color space. Then pixels are grouped iteratively based on their closest `seeds', which are the initial values for superpixel centers. Pixel grouping is done according to both the chromatic similarity $D_{lab}$ and spatial proximity $D_{spatial}$ of the pixel with the associated seed. The metric used to quantify this distance, $D_{total}$ is given as
\begin{align*}
    D_{total}&= D_{lab} + \theta* D_{spatial},\; \mbox{where}\\ 
    D_{lab}&= \left \| (l,a,b)_{i^{th} pixel} - (l,a,b)_{k^{th}seed} \right \|\\
    D_{spatial}&= \left \| (x,y)_{i^{th} pixel} - (x,y)_{k^{th}seed} \right \|
\end{align*}
for the $i^{th}$ pixel and the $k^{th}$ seed. $\left \|.  \right \|$ is a Euclidean distance and $\theta$ is defined by $\theta = 1/c^2$ where $c$ is a user-defined \textit{compacity} parameter. The seeds value is propagated temporally, to generate an output of average RGB color value per temporal superpixel per frame.

 Let $C_p(t)$ denote the 3-dimensional per-channel average YUV color signal of a superpixel $p$ in frame $t$. The 3 average YUV values and grouping all $K$ superpixels gives a 2d matrix of size $K\times 3$. Stacking these over a temporal window of $T$ frames, we obtain a $3\times K\times T$ tensor, which can be seen as a $K\times T$ color image which we call a \textit{spatial-temporal map}. This map is 10 seconds long, while each clip lasts around 1 minute. We slide a temporal window over the clip to obtain M maps per clip using a  window of 10 seconds (stride = 0.5s). \vspace{-3mm}

\section{Experiments}
\label{sec:exp}

We compare against two state of the art rPPG methods in HR-CNN \cite{vspetlik2018visual} and RhythmNet \cite{niu2018vipl}. Both these methods involve pre-processing the input video frames using face-detection (with an additional alignment step in RhythmNet), cropping and re-sizing using specific toolboxes. 
We use train-test splits as specified by the authors.  For our dataset, we make use of a train-test split featuring 8 subjects in the train set and 3 in the test set. The number of superpixels is $K = 300$. 

\begin{figure*}
    \centering
    \vspace{-5mm}
    \includegraphics[width=0.9\linewidth, trim=0 0 0 20,clip]{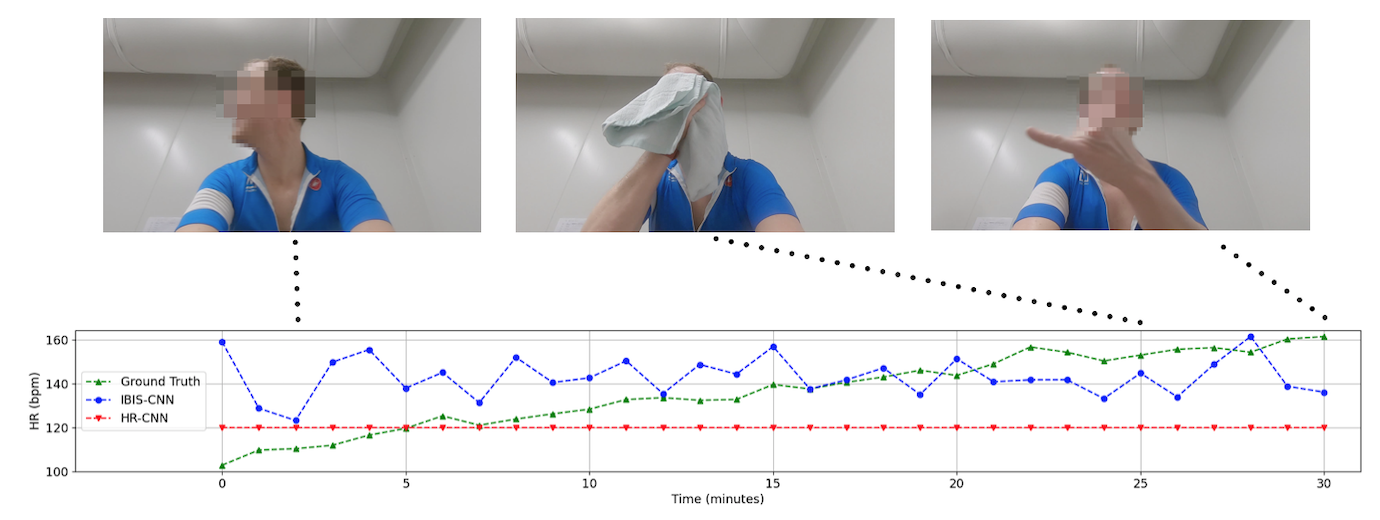}
    \vspace{-0.5cm}
    \caption{Predictions of the IBIS-CNN model compared with he ground truth and HR-CNN on subjects from the test set. The HR-CNN model outputs a constant value, while our predictions exhibit learning behavior, following the ground truth. \vspace{-0.5cm}}
    \label{fig:predictions}
\end{figure*}


\begin{table}[]
\centering
\resizebox{\columnwidth}{!}{%
\begin{tabular}{@{}llccc@{}}
\toprule
                      & Model           & PURE           & ECG-Fitness    & IntensePhysio  \\ \midrule
\multirow{3}{*}{RMSE} & HR-CNN          & \textbf{11.00} & 19.15          & 25.27          \\ \cmidrule(l){2-5} 
                      & RhythmNet       & 19.67          & 20.47          & 32.67          \\ \cmidrule(l){2-5} 
                      & IBIS-CNN (ours) & 11.99          & \textbf{17.03} & \textbf{22.01} \\ \midrule
\multirow{3}{*}{MAE}  & HR-CNN          & \textbf{8.72}  & 14.48          & 22.19          \\ \cmidrule(l){2-5} 
                      & RhythmNet       & 17.46          & 16.82          & 28.36          \\ \cmidrule(l){2-5} 
                      & IBIS-CNN (ours) & 9.39           & \textbf{13.75} & \textbf{16.53} \\ \bottomrule
\end{tabular}%
}
\caption{RMSE and MAE results for the proposed IBIS-CNN and two baseline methods in HR-CNN and RhythmNet on the test sets of PURE, ECG-fitness and IntensePhysio (ours). \vspace{-6mm}}
\label{tab:results}
\end{table}

Due to lack of publicly available code for RhythmNet \cite{niu2018vipl}, we re-implemented it as per the authors description in their paper. The HR-CNN \cite{vspetlik2018visual} model and code is publicly available. We evaluate with mean absolute error (MAE) and root mean square error (RMSE) calculated over the number of spatio-temporal maps. Results in Table ~\ref{tab:results} 
show that our method achieves similar performance to others on PURE and ECG-Fitness, validating our baseline. Yet, our baseline significantly outperforms others on our new IntensePhysio dataset.


All models show poor performance on our dataset. To establish the challenging nature of the IntensePhysio dataset for methods that rely on face tracking, we run an off the shelf face detector (Dlib \cite{dlib09}) and find that \emph{on average we are able to detect faces in  $\sim$28$\%$ of frames per video in our dataset.On the other hand we observed a successful face detection in $\sim$74$\%$ frames per video in ECG-Fitness}. Since the face region (particularly the cheeks and forehead) contain the most information on heart rate \cite{chen2018deepphys}\cite{vspetlik2018visual}, this leads to a poor performance when models rely on face detection and tracking. 

The predictions of our IBIS-CNN model and HR-CNN along with the ground truth are shown in Fig. \ref{fig:predictions}. A few sample frames from the video are also presented with the dotted lines connecting them to the minute of extraction. In these frames, there is a lot of specular reflection on his skin (due to sweating), the subject's  face is not visible. This happens particularly around the $30^{th}$ minute, where the error is larger. The HR-CNN model outputs a constant value, indicating that it might not be learning relevant rPPG features. The IBIS-CNN model is able to predict heart rate reasonably well despite occlusions, specular reflections etc. However, we also note that the predictions from IBIS-CNN are not always correlated with the ground truth, especially in the second plot of \ref{fig:predictions}. This could be because of rapid motion and would be interesting future direction of research. IBIS-CNN does not solve all the challenges posed by the IntensePhysio dataset but is an alternative baseline with a new approach.

\textbf{Comparison of input representations.} We compare two existing methods for temporal superpixel generation - IBIS~\cite{bobbiaibis} and TS-PPM~\cite{lee2017temporal}.
We find that the performance of TS-PPM is lower than that of IBIS-CNN (as seen in Table 3 possibly because, in the IBIS method pixel membership is constrained to not vary beyond a certain threshold to maintain temporally coherent RGB traces.

As a pre-processing step, the IBIS superpixel generation is more computationally efficient than the TS-PPM. To generate the superpixel results, IBIS processes $5.33$ frames/sec on average whereas TS-PPM was averages $0.56$ frames/sec. Thus, the IBIS method is more suited to the rPPG estimation task. These results were obtained using PURE dataset.

\begin{table}[htp!]
\centering
\resizebox{\columnwidth}{!}{%
\begin{tabular}{@{}ccc@{}}
\toprule
                      & Model           & Validation error for PURE \\ \midrule
\multirow{2}{*}{RMSE} & TS-PPM \cite{lee2017temporal}         & 13.97                     \\ \cmidrule(l){2-3} 
                      & IBIS-CNN (ours) & \textbf{11.99}                     \\ \midrule
\multirow{2}{*}{MAE}  & TS-PPM \cite{lee2017temporal}     & 10.81                     \\ \cmidrule(l){2-3} 
                      & IBIS-CNN (ours)  & \textbf{9.39}                      \\ \bottomrule
\end{tabular}%
}
\caption{Comparison between temporal superpixel methods for heart rate estimation.}
\vspace{-5mm}
\end{table}

\section{Conclusions}
\label{sec:conclusions}
We present IntensePhysio, a challenging new dataset for heart rate estimation. The dataset features large subject motion with frequent face occlusions and cases of facial region absent from the frame entirely. Through a comparative study,  we observe a considerable degradation in the performance of the existing state of the art methods on this new dataset, especially methods relying on face detection and tracking. This highlights IntensePhysio as a challenging dataset for heart rate estimation, indicating that occlusion and non-visible facial regions are key factors for this performance degradation. Hence, we propose IBIS-CNN as a new baseline method for heart rate estimation (using temporal superpixels) which significantly outperforms state of the art methods on our challenging new dataset in addition to the existing ones. 
However, the IBIS-CNN baseline predictions are not strongly correlated with the actual ground truth always.This shows that there are further problems posed by our dataset that require addressing. Also, there is a need for a thorough hyperparameter tuning while generating these temporal superpixels using IBIS, specific to our task. So, it would be of significance to investigate the possibility of developing IBIS-CNN as an end to end learnable pipeline.

\bibliographystyle{IEEEbib}
\bibliography{strings,refs}

\end{document}